# Predicting Energy Consumption of Ground Robots On Uneven Terrains

Minghan Wei, and Volkan Isler

***Abstract*—Optimizing energy consumption for robot navigation in fields requires energy-cost maps. However, obtaining such a map is still challenging, especially for large, uneven terrains. Physics-based energy models work for uniform, flat surfaces but do not generalize well to these terrains. Furthermore, slopes make the energy consumption at every location directional and add to the complexity of data collection and energy prediction. In this paper, we address these challenges in a data-driven manner. We consider a function which takes terrain geometry and robot motion direction as input and outputs expected energy consumption. The function is represented as a ResNet-based neural network whose parameters are learned from field-collected data. The prediction accuracy of our method is within 12% of the ground truth in our test environments that are unseen during training. We compare our method to a baseline method in the literature: a method using a basic physics-based model. We demonstrate that our method significantly outperforms it by more than 10% measured by the prediction error. More importantly, our method generalizes better when applied to test data from new environments with various slope angles and navigation directions.**

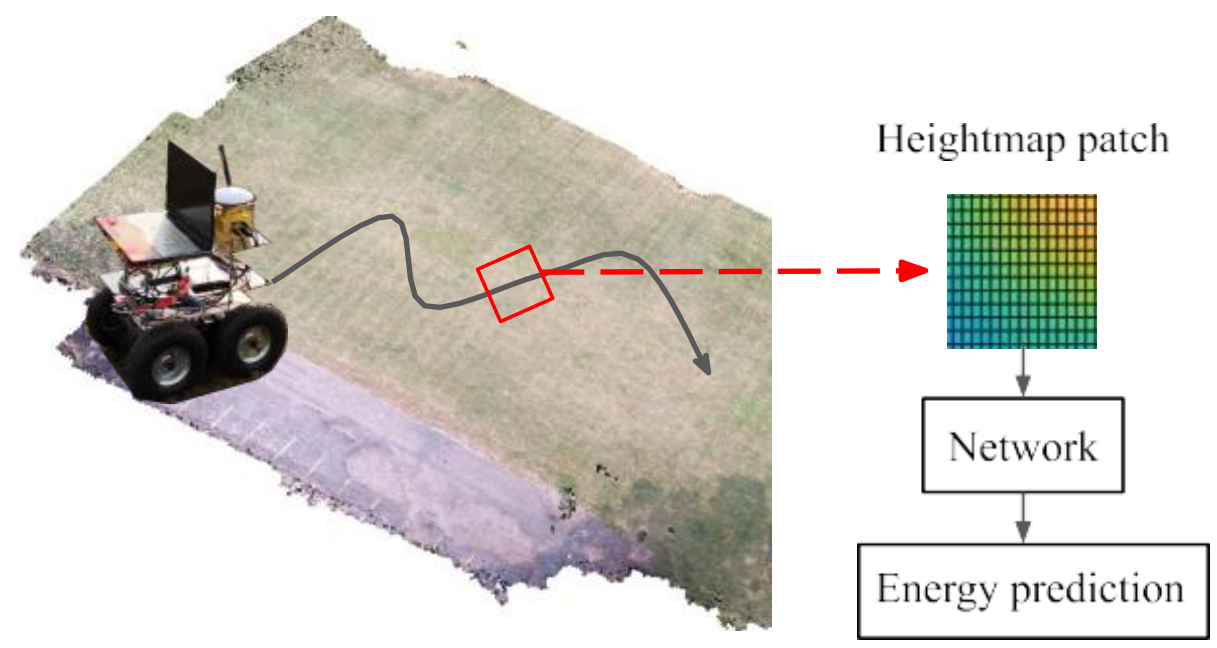

Fig. 1: The left part of this figure shows one view of the data collection and test sites. We predict the energy consumption of a ground robot when it navigates on an off-road environment with slopes. The prediction is based on the path segments. For each segment, we extract the heightmap patch and use a network to predict the energy consumption for navigating across this patch. The energy cost of the entire path can be calculated by summing up the costs of all the segments.

## I. Introduction

Autonomous navigation is a crucial capability for field robots to perform tasks such as environment monitoring, searching, and delivery. Many of these robots are powered by batteries and have limited energy budgets. Meanwhile, it has been shown that actuation is one of the main consumers of the energy on a robot [22] [16]. Therefore, planning energy-efficient paths is important for field robots.

Path planning is a fundamental robotics research topic and has been widely studied, including variations focusing on energy efficiency. Planning algorithms, such as Dijkstra, $A^*$, and Rapidly-exploring random tree($RRT$), usually require a map of the planning space. To apply existing planning algorithms to find the energy-optimal path to a goal location, an energy-cost map of the given environment is required. From the energy-cost map, we can also check whether a given path can be followed with the remaining energy budget of the robot. However, there is a disconnect between current theoretical results and practical applications due to the difficulty in obtaining energy-cost maps for energy-efficient path planning.

To build energy-cost maps of given environments, some work in the literature estimates robot energy consumption using standard physics-based models based on friction and gravity force [14], [21], [22]. This method does not generalize to uneven terrains. Our previous work addresses this issue using data-driven approaches [25]–[27]. We learn from aerial images and ground energy consumption measurements to efficiently build energy-cost maps for environments without slopes. In this paper, we study the energy-consumption prediction problem for general environments with slopes.

On uneven terrains, the energy consumption at a location can be directional because of the varied slope angles along navigation directions. Two factors can affect the energy consumption of a ground robot when following a path $L$. (1) The path length of $L$. Usually the longer the path is, the higher energy consumption it takes. (2) The slope angles along $L$. For example, larger slope angles uphill costs more energy for climbing. Therefore, if we directly predict the energy consumption of a path using a learning-based method, we may have to collect a large number of paths with varied length and slope angles for training, which is not an easy task. Alternatively, we can just predict the energy consumption for moving from a single location for a unit distance (e.g. one meter). The navigation direction at this location is continuous and we need to predict the energy consumption for continuous navigation angles. A simple solution is to discretize the navigation direction to $n$ parts. A large value for $n$ makes the data collection tedious at each location, while a small value negatively affects the prediction accuracy.

In this paper, we address the problem of predicting the energy consumption of a ground robot on uneven terrains

[1,2]The authors are with the Department of Computer Science and Engineering, University of Minnesota, 200 Union Street SE, Minneapolis, MN, US, 55108. weixx526, isler@umn.edu. This work is funded in part by NSF grant #1525045, NSF grant #1617718, NSF grant #1849107, and MN State LCCMR program.

with slopes. Our method is based on predicting the energy consumption of a path segment from the heightmap patch to resolve the challenges, as shown in Fig. 1. The heightmap patch is fed into a ResNet-based network to learn the energy consumption for the ground robot to navigate across this patch. The energy consumption of the entire path can be calculated by summing up the costs for all the segments.

Our main contributions in this paper are: (1) We present a learning-based method for energy prediction that can resolve the challenges on data collection, network training, and generalization. The network is trained with data collected at one site and then applied to new areas for energy prediction. We empirically validate that our method generalizes well to navigation directions, slope angles within $\pm 5°$, and environments of the same terrain class that are not included for training. (2) We also show that when applied to a new terrain class that is unseen during the training, the prediction results are still useful for planning energy-efficient paths. (3) We compare our method with a representative baseline method that uses a standard physics-based model and show that our network outperforms them by more than 10% measured by the prediction error.

## II. Related Work

Field applications, such as precision agriculture, environmental monitoring, search, and mapping, require us to plan paths for robots to navigate in the working space. Since robots are subject to energy constraints in fields, how to make robots navigate in an energy-efficient manner is gaining increasing research attention.

Many planning algorithms need an environment map to find a path. In the case of energy-efficient path planning, energy-optimal paths between any pair of locations can be calculated using Dijkstra's algorithm [6] or $A^*$ [10] assuming that worlds are represented by a grid and an energy-cost map is available. RRT algorithm [13], [17] or its variants [1], [12] can be applied to trade off between the planning efficiency and path optimality. In online path planning where an environment map is not available, the robot needs to explore the environment. For example, the $D^*$ algorithm is used for the case where the map is not fully observed and can change during navigation [11]. We refer the readers to surveys [8], [9], [15] for additional path planning methods.

However, efficiently building an energy-cost map is not an easy task. The inconvenience of measuring the friction coefficients everywhere on non-uniform and rough terrains, especially for large environments, undermines the applicability of physics-law-based methods [14], [21], [22]. To efficiently acquire friction coefficients over the entire environment, Quan et al. [18] fit the friction coefficients based on the current energy measurements for the measured areas first, and then estimate the coefficients for the rest area using the Gaussian Process method. This method works when available measurements scatter across the environment. As we show in Sec. V-D, this method does not generalize well to new environments with various navigation directions and slope angles, especially when there are no measurements available in the test fields.

Other factors that affect ground robot energy consumption include robot acceleration and angular velocity. Tokekar et al. presented a solution to finding optimal paths and velocity profiles for car-like robots to minimize the energy consumed during motion [24]. The work in [7] studies the energy consumption of a skid-steered to optimize the turning cost. Tiwari et al. [23] present a unified model which combines several forces that consume robot energy for aerial, ground, and marine vehicles to estimate the operational range. In our work, we would like to not only estimate the robot energy consumption, but also be able to plan energy-optimal paths to goal locations by predicting an energy-cost map of given environments.

Visual inputs are useful for building environment maps. For example, ground traversability can be estimated from images [3], [20] and guides ground robots to follow safe paths. An aerial vehicle can explore an environment faster than a ground robot. Therefore, it can help the ground robot navigate to goal locations faster by providing map information from aerial images [5].

In our previous work, we have presented that for environments without slopes, which means the energy consumption at a location is the same for all navigation directions, the energy-cost maps can be efficiently built from aerial images and ground robot energy measurements [25]–[27]. On general terrains with slopes, the energy consumption at a location is directional, which makes our previous method not applicable. In this paper, we study the effects of slopes and extend our previous results for predicting the energy consumption of ground robots in general fields.

## III. Problem Formulation

Before formally stating our problem, we first introduce important concepts and notations.

A *heightmap* $H$ of an environment is a function that takes a 2D coordinate as the input and returns the altitude for this position. The robot we work with can measure its energy consumption during navigation by the battery voltage $V$ and current $I$. Let $p$ and $q$ be two locations on the robot trajectory. The energy cost for navigating from $p$ to $q$ is calculated by $E = \int_{t=0}^{t=t_q} V(t)I(t)dt$, where $t$ is the time. The robot is at $p$ when $t = 0$ and at $q$ when $t = t_q$. We use $F$ to denote a function that takes $H$ and a path $P$ as input and outputs the energy consumption of the ground robot for following $P$.

With the above information for the environment heightmap and the robot energy measurements, we can formulate the problem. We are given a ground robot that can measure its energy consumption as it moves. The ground robot navigates in an off-road environment whose heightmap $H$ is available. Given any path $P$ in this environment, our goal is to learn a prediction function

$$\hat{E} = F(H, P) \tag{1}$$

so that the difference between $\hat{E}$ and $E^*$, $(\hat{E} - E^*)^2$, is minimized, where $E^*$ is the ground truth energy cost of the robot for following $P$.

The terrain surface in fields is usually uneven, which makes the energy consumption non-uniform. Meanwhile, the slopes make energy consumption directional at most locations. In this paper, we first study the energy prediction function $F(H, P)$ for environments of the same terrain class (grass, as shown in Sec. IV). In Sec. V-E, we present that the learning results are useful for planning energy-efficient paths on a new terrain class (dirt).

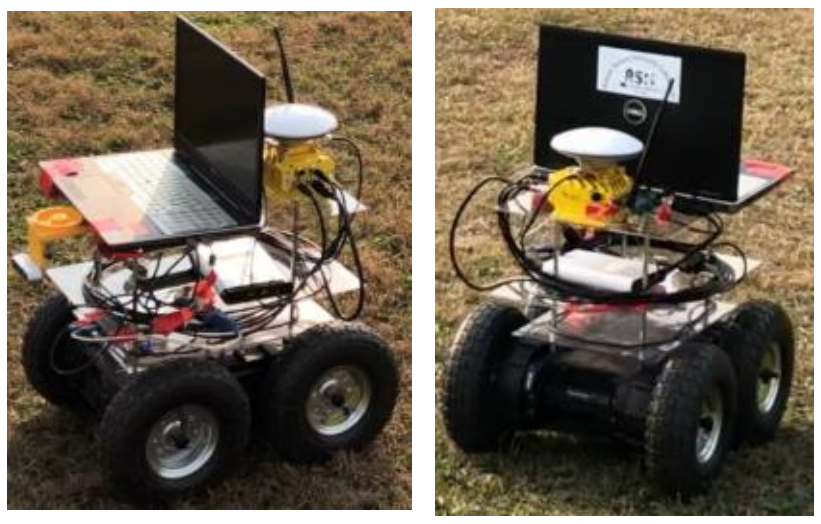

Fig. 2: The ground robot we use for data collection.

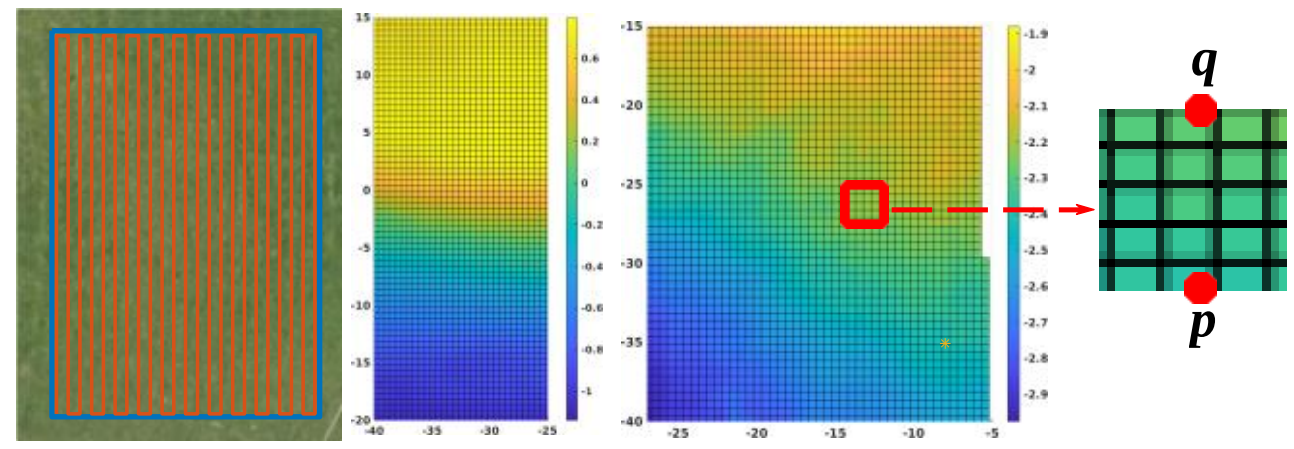

Fig. 3: Data collection and processing for training. From left to right. (1) We use the ground robot to follow a lawnmower (back-and-forth) motion pattern to cover this area and collect the measurements. (2), (3) The global heightmap for two grass areas in our work. The origin is the RTK GPS base station. (4) We extract heightmap patches along the robot trajectory as the network input. The network is trained to predict the energy consumption for navigating across this patch.

## IV. Energy Prediction based on Heightmap Patches

In this section, we demonstrate our method to learn the prediction function $F(H, P)$, where $H$ is the height map of the navigation environment and $P$ is the robot path. We start by introducing the data collection and processing procedures. We then present the network architecture and training. The test results on the data collected near the training environment are also presented.

### A. Ground Robot and Data Collection Process

In this subsection, we introduce the ground robot used in this paper and the data collection procedure. We also describe how we process the data to train a network for energy prediction.

The four-wheel-drive ground robot we use is from Rover Robotics (*4WDRoverPro*). It is powered by a lithium-ion battery whose nominal voltage is 28.8 volts. Fig. 2 shows two views of the system. The ground robot is able to read its battery voltage and current to measure the energy consumption for navigation. All other onboard devices such as the GPS and the laptop are powered by a separate source.

In this paper, we consider only navigable surfaces and therefore do not consider situations where the robot sinks or gets stuck, for example, into sand or a mud puddle. In these cases, the robot would spend all its energy without moving forward.

We use the robot platform in Fig. 2 to collect energy consumption data on a grass area of a park first. The total size of the environment is around $60m \times 40m$. We guide the ground robot to navigate along a trajectory at a constant speed for data collection. During the motion, the robot records its 3D position, the battery voltage, current, and the system time at $10Hz$. For the robot position, we use the RTK GPS from *SwiftNav*, which can provide 3D coordinates of centimeter-level accuracy. The GPS is rigidly fixed to the ground robot so that the position change of the GPS is the same as that of the robot. The voltage, current, and time can be used to calculate the energy consumption of the robot for any part of the trajectory, as mentioned in Sec. III. Fig. 3 shows one field for data collection in our work. The slope angle is between $-5°$ and $5°$. The robot follows a lawnmower (back-and-forth) motion pattern [4] at a speed of $0.5m/s$.

### B. Heightmap-Patch-based Energy Prediction

There are two primary challenges when learning to predict the cost of an arbitrary path: (1) Learning to predict the energy cost of the entire path requires collecting a large number of example (training) paths with varying lengths, slope angles, and robot navigation directions. It adds to the difficulty of data collection. (2) How to input paths of varied lengths, slope angles, and navigation directions into a network may require extra efforts on network architecture design.

To resolve this issue, we divide a path into segments of unit length, and then estimate the energy consumption for each segment. We sum the costs of all the segments to calculate the energy consumption of the entire path, as detailed below.

We divide a path and extract the training data from the raw measurements as follows: Let $p = (p_x, p_y, p_z)$ and $q = (q_x, q_y, q_z)$ be two positions on the robot trajectory where the robot moves from $p$ to $q$. The distance between $p$ and $q$ when projected onto the horizontal plane is $1m$. That is: $(p_x - q_x)^2 + (p_y - q_y)^2 = 1$. We extract the heightmap patch of size $1m \times 1m$ that has $p$ as the middle point on one edge, and has $q$ as the middle point on the opposite, as shown by the square in the third and forth subfigure of Fig. 3. We represent this heightmap patch as a $256 \times 256$ matrix $M$. $M$'s first row matches the edge of $q$ and the last row is for the edge of $p$. We subtract $M$ by its minimum value so that the lowest positions in $M$ have height values of zero. Each value in $M$ represents the height of the corresponding position. We use $M$ as the network input for the training data. Note that we use the heightmap patch $M$ instead of the height values of the path segment $pq$, which can be represented as a vector, to learn the energy consumption. It is because the robot has a certain width and the energy consumption is not only affected by the trajectory of the robot center, but also nearby areas where the left and right wheels pass. Meanwhile, using $M$ allows us to take advantage of existing network architectures and the pretrained parameters, as described in Sec. IV-C.

Let $E$ be the energy consumption for navigating from $p$ to $q$. Viewing from $M$ , $E$ is the energy consumption of the ground robot for moving along the centerline in $M$ from bottom to top.

The energy prediction function $F(H, P)$ takes a path $P$ and the corresponding heightmap $H$ as inputs. After the above division, the problem in Sec. III is equivalent to the version as follows: For any given path $P$, we divide $P$ into the segments, $P = \{P_1, P_2, ..., P_n\}$ and extract the heightmap patches $M = \{M_1, M_2, ..., M_n\}$ from the global heightmap H. The ground truth energy consumption for each segment is also available: $E^* = \{E^*_1, E^*_2 ..., E^*_n\}$. Our goal is to learn a function:

$$\hat{E}_i = f(M_i, P_i)$$
$$min: \sum_{i=1}^{n} |E_i^* - \hat{E}_i|^2 \tag{2}$$

Here the function $f(M_i, P_i)$ takes a path segment with the heightmap patch as the input, and outputs the corresponding energy consumption for it.

We can now focus on training a network to predict the energy consumption given a heightmap patch, since in each patch $M_i$, $P_i$ is fixed. We use $e_i^* = \frac{E_i^*}{V}$ as the ground truth for network training to improve the scalability, where $V = 28.8$ which is the battery nominal voltage value. Similarly, the network output should be $\hat{e_i} = \frac{E_i}{V}$.

We separate the data from the environment in Fig. 3 for training and testing. We use the patches from the left one-third area for training, and the rest for testing. We also collect data from other environments to test the generalization capability of our method, as shown in Sec. V.

In the problem formulation, we suppose that the global terrain heightmap $H$ is given for extracting local height map patches. There are a variety of ways to obtain this heightmap in a practical situation. For example, we can build a 3D reconstruction from aerial images of the field when the altitude of the drone is known. In our work, since we have used the ground robot to cover the entire field to collect training and test data, we fit a surface with the GPS positions along the trajectory. In this way, we do not need additional steps for acquiring the global height map.

### C. Network Architecture and Training Details

Our network input is a heightmap patch, represented by a 256 × 256 matrix, and the output is a single value for energy consumption for navigating across this patch. We use the 'ResNet18' network architecture pretrained on 'ImageNet' for this purpose. We remove the 'ResNet18' prediction head and add three fully-connected layers with the output sizes of 512, 256, 1. Each of these layers has an 'ReLu' activation function except the last one.

The pretrained 'ResNet18' on 'ImageNet' is for image classification, which is different from our energy prediction task. We compare three training strategies: (1) Fixing the pretrained weights in 'ResNet18' and only train the top fully-connected layers; (2) Training the network from scratch. (3) Using the pretrained weights in 'ResNet' and fine-tune it as well as the top fully-connected layers. We find that the third option works best in terms of prediction accuracy and training time. The loss function for the final layer is mean square loss (MSE). The learning rate is 0.0001. The data size from the environment in Fig. 3 is around 4400 (4400 heightmap patches and the corresponding energy measurements), and we use one-third of them for training. The training process takes around one hour with an Nvidia Quadro P3200 GPU.

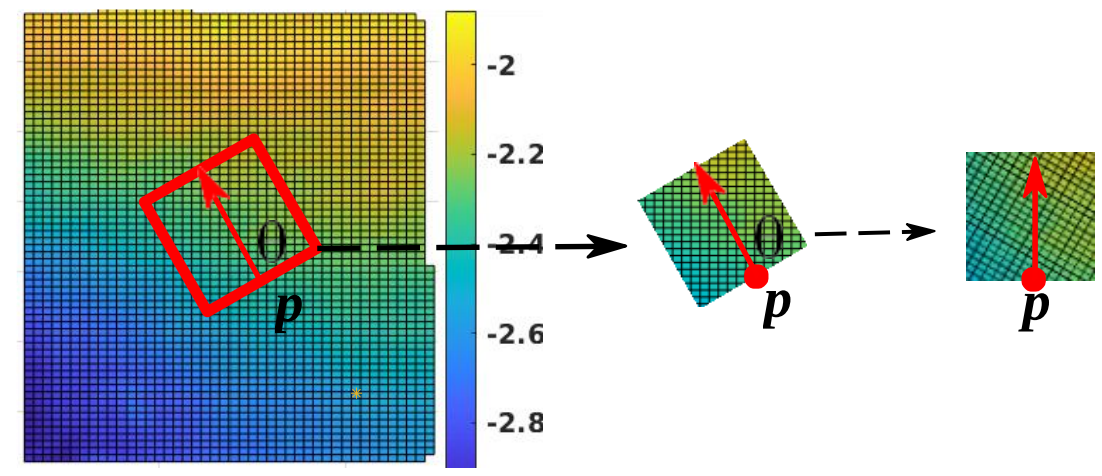

Fig. 4: When the robot navigates from $p$ along $\theta°$ direction, we use the square region that has $p$ at the bottom center and orients by $\theta°$ from the north to extract the heightmap patch. When represented as a matrix $M$ , it is still the same as the case in Fig. 3 that the robot navigates along the centerline of $M$ from the bottom to the top.

### D. Prediction for Continuous Navigation Directions

We have used a back-and-forth motion pattern for data collection in Fig. 3, where the navigation direction corresponds to the north and south. In this case, we use a rectilinear square region to extract the heightmap patches. However, in a practical scenario, the robot can move in any direction. In this subsection, we present additional details on extracting heightmap patches so that our method can be applied to predict the energy consumption for continuous navigation directions at a location.

Let $p$ be a position in the environment and the robot navigates along the $\theta°$ direction, where 0° is the north direction. As has been introduced, the heightmap patch is represented as a matrix, and our ResNet-based network predicts the robot energy consumption for moving along the centerline of the matrix from the bottom to the top. To extract such a matrix for any $p$ and $\theta$, we first use a square region, which has $p$ as the middle point at the bottom edge and oriented by $\theta°$ from the north, to obtain the heightmap patch, as shown in the left part Fig. 4. We then record the height values in this patch as a matrix $M$ , as shown by the rightmost part in Fig. 4. In this way, using $M$ as the input, the network predicts the energy consumption at $p$ for moving along the $\theta°$ direction.

### E. Practical Deployment Procedures

In this subsection, we summarize how our heightmap-patch-based method can be deployed for energy mapping and energy-efficient path planning in a practical setting. A typical application scenario is that a ground robot needs to navigate in an agriculture field for tasks such as inspection and harvesting. Our method can build an energy cost map for the field so that the robot can follow the energy optimal paths to goal locations.

The first step of our method is to build a global heightmap for the given environment. One of the fastest ways can be to use an aerial vehicle (UAV) equipped with sensors such as a camera or Lidar to cover the area, which is readily available on the market. The sensors can record the altitude of each location in the global heightmap. There is also dedicated work on how to reconstruct surfaces from sensor information [2].

The next step is to extract the heightmap patches as the inputs for a pretrained network to predict the energy cost. The terrain class of the testing area may have a different energy profile as the training environment. In Sec. V-E, we demonstrate how to improve the prediction accuracy for unseen environments during training.

## V. Performance Evaluation

In this section, we evaluate the performance of our method in four steps: (1) We first show the test results with the data from the back-and-forth motion pattern (Sec. V-A). (2) We then demonstrate that the network trained with a limited set of navigation directions and slope angles (see Fig. 3) can accurately predict the energy cost for other moving directions and slope angles in Sec. V-B. (3) We test the network performance on a new grass area and show that our method generalizes well to unseen environments of the same terrain class in Sec. V-C. (4) Lastly, in Sec. V-E, we apply the trained network to a new terrain class (dirt), and present that the predicted energy values by our network can still be useful for planning energy-efficient paths to goal locations.

To show the advantage of our method of learning the energy consumption from heightmap patches, we compare it to a physics-based model to show that our method significantly outperforms it by around 10% and has a better generalization capability in Sec. V-D.

### A. Test Results on Data from Back-and-Forth Motion

We first demonstrate the test results on the data collected in Fig. 3. We plot the prediction and ground truth measurements in Fig. 5. The $x$-axis corresponds to the heightmap patches along the robot trajectory and the $y$-axis is the corresponding scaled energy consumption $\hat{e}$. As mentioned in Sec. IV-B, all the network outputs (predictions) are scaled by the battery nominal voltage value 28.8. Recall that we split the data and only use the left one-third area for training and the right two-thirds area for testing.

We quantify the difference between the prediction ($\hat{e}$) and the ground truth ($e^*$) by the relative error:

$$r = \frac{abs(\hat{e} - e^*)}{e^*}. \quad (3)$$

The average error rate of our prediction in this environment is 9.1%. Note that the ground truth energy measurements by themselves are noisy. It is not expected to have a perfect prediction. Our test results in Fig.5 reflect the energy consumption difference under different ground conditions. A detailed performance comparison with representative baseline methods is presented in Sec. V-D.

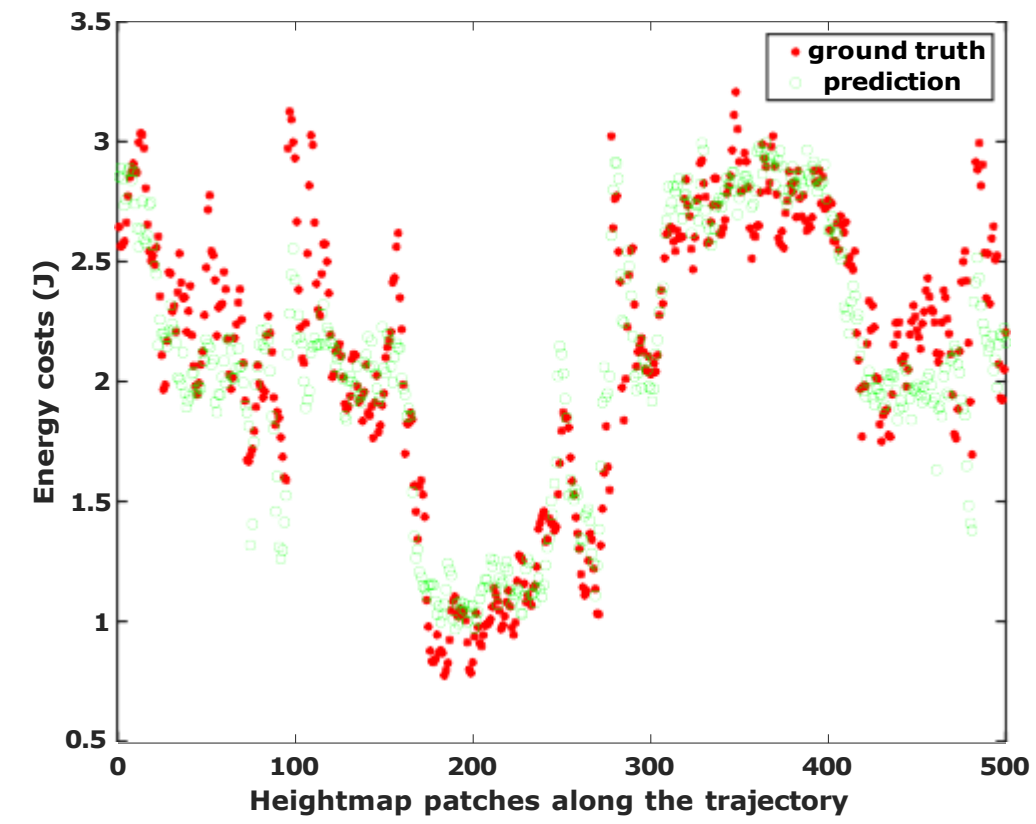

Fig. 5: Part of the prediction results on the test data from the area in Fig. 3. The x-axis is the heightmap patches extracted along the trajectory. The values for y-axis are the energy consumption values in Joule (J) scaled by 28.8 (the nominal voltage of the robot battery). The x, y-axis meaning is the same for Fig. 6, 7, 8, 10, 11. The average error is 9.1%.

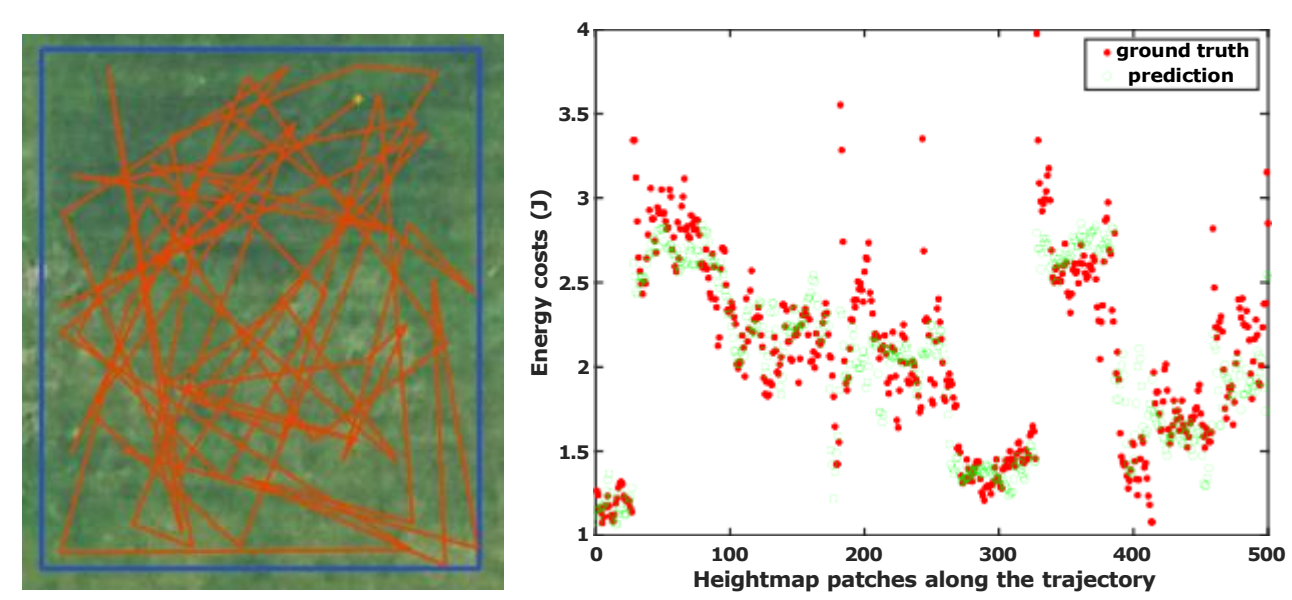

Fig. 6: We collect data by guiding the ground robot to follow randomly-selected waypoints. Thus the test data contains multiple navigation directions and slope angles that are not included in the training. We plot 500 of the test results, with the ground truth energy consumption and the prediction of our network on the right. The average error rate is 10.9% and similar to the test in Fig. 5.

### B. Generalization to Continuous Navigation Directions and Slope Angles

To show that our ResNet-based network generalizes for continuous navigation directions and new slope angles, we randomly sample 60 waypoints in the training environment and let the robot reach them one by one. Since the waypoints are selected randomly, the navigation directions and slope angles also vary from the training data. Fig. 6 illustrates the robot trajectory. We extract heightmap patches and the corresponding energy consumption along the trajectory for testing.

We plot part of the testing results in the right subfigure of Fig. 6. The energy prediction deviates from the ground truth with an average error of 10.9%. This experiment verifies that our method generalizes to new navigation directions and slope angles in the surrounding areas, since the error is similar to the test in Fig. 5 where testing data are collected using the same motion pattern as the training.

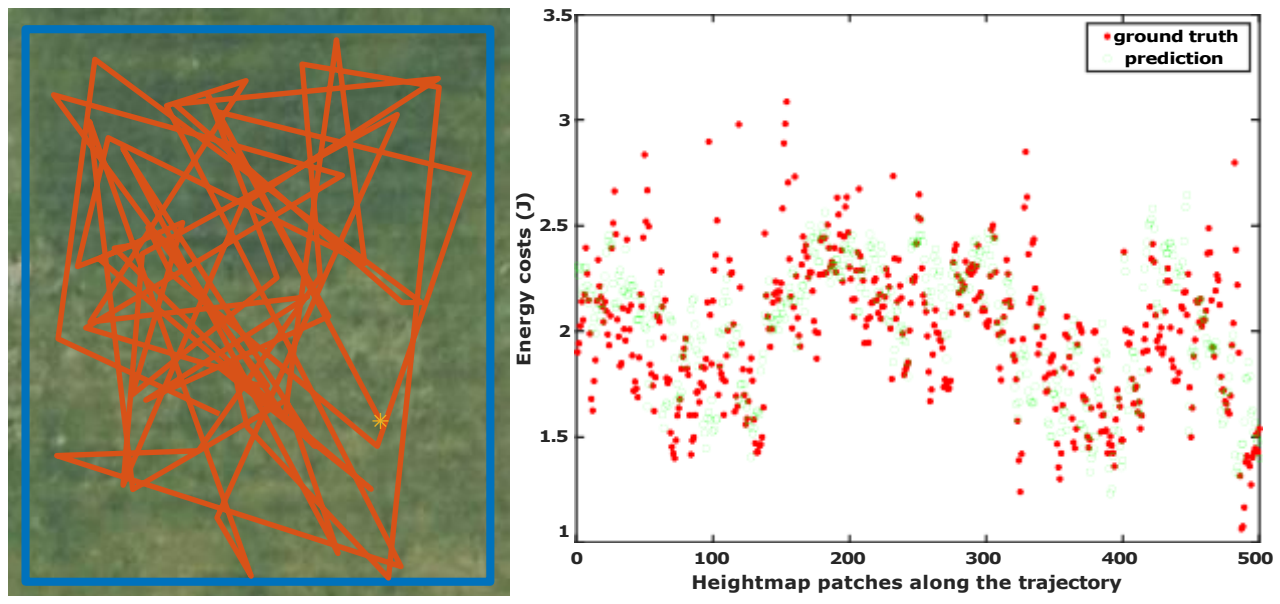

Fig. 7: To further show the generalization capability of our method, we collect new test data at another grass site. By following random waypoints (left subfigure), the test data includes varied navigation directions and slope angles. We plot part of the energy consumption estimation and ground truth on the right subfigure. The predicted value follows the trend of the ground truth. The average error rate is 12.9%.

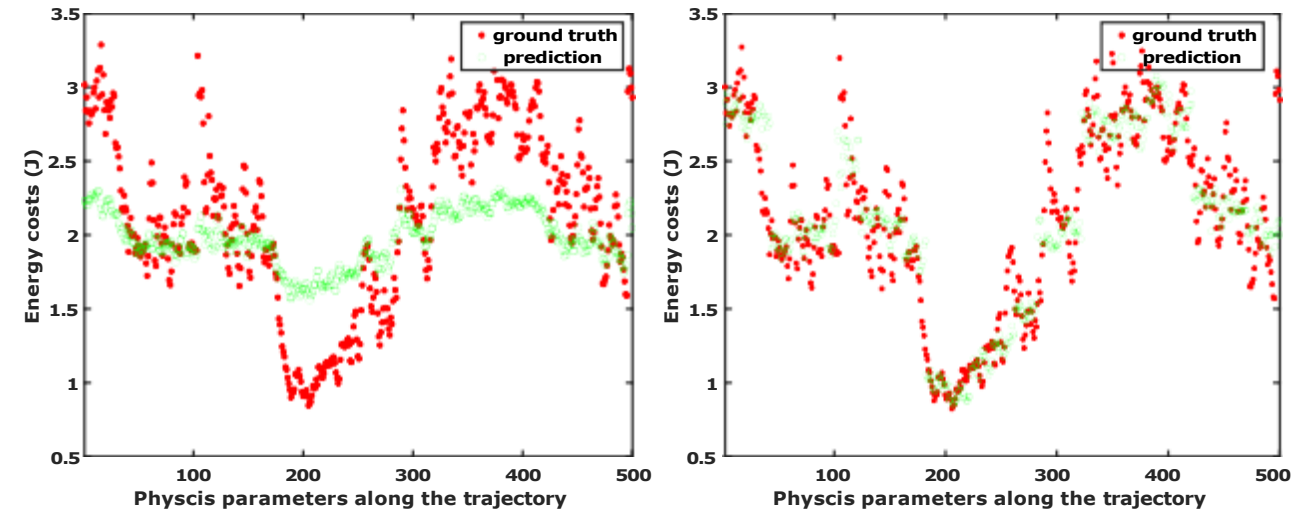

Fig. 8: Part of the estimation results from the model using standard physical laws based on friction and gravity force. *left.* When training and testing data are from separate areas, the average estimation error is 23.6%. *right.* When training and testing data are five-meter apart, the error improves to 10.8%. Compared to Fig. 5, we can see that the learning-based method has higher accuracy especially when training and testing data are from separate areas.

### C. Generalization to New Environments

To demonstrate the energy prediction performance in unseen environments for training, we use the ground robot to take energy measurements at another grass site, which is around $2km$ away from the grass area for training.

Similar to the previous subsection, We guide the robot to navigate to a set of randomly selected waypoints so that the test data contains various navigation directions and slope angles, as shown on the left of Fig. 7. The network estimation and the ground truth are plotted on the right. The prediction still follows the trend of the ground truth measurements along the trajectory. The overall error rate is 12.9% and close to the test in Sec. V-B. In the next subsection, we compare it with a physics-based model to show the better generalization capability of our method.

### D. Comparison with a Model-based Approach

To demonstrate the advantage of our method, we compare it with an approach using a simple physics-based model. We present that our method has a higher prediction accuracy as well as better generalization capability in new environments. In an ideal situation, a ground robot's energy consumption for navigating at a fixed speed $v$ can be calculated by [18], [19], [21], [22]:

$$E = \int (\mu(t) mg \cos(\theta(t)) + mg \sin(\theta(t))) v dt, \tag{4}$$

where $\mu$ is the friction coefficient (or rolling resistance coefficient), $m$ is the robot mass, $g$ is the gravitational acceleration constant, and $\theta$ is the slope angle on the navigation direction. This equation mainly considers the effects of friction force and gravity force. The term $vdt$ is the navigation distance and can be obtained by GPS. Meanwhile, $m$, $\theta$, $v$, and $g$ are also available. The only unknown in the equation is the friction coefficient $\mu$. Measuring $\mu$'s value for all the locations and navigation directions is not a feasible solution on rough terrains. However, we can calculate the values of $\mu$ for the locations in the training data where we have the energy measurements using Equation 4 as follows.

Assume the $\mu$ does not change over a path segment of length $d$ ($d = 1m$ in our work). Then based on Equation 4, we have:

$$E = (\mu mg \cos(\theta) + mg \sin(\theta)) d \tag{5}$$

In Equation 5, $\mu$ is the only unknown and thus can be directly calculated. When we have multiple measurements (multiple equations in the form of Equation 5, a least-square method can be used to fit a value for $\mu$ to minimize the error.)

We use the measurements in Fig. 3 for the performance comparison. Depending on how we split the training and testing data, we present two cases. (1) When the training and testing data are from separate areas (left one-third and right two-thirds on the same terrain class), we first use 'Least Square' to fit a friction coefficient with all the training data. We apply this value as the global friction coefficient to the testing data and plot the prediction in the left subfigure of Fig. 8. We could see that in this case, the friction coefficient value fitted by Equation 4 does not yield accurate results. The average error rate is 23.6%. (2) We study the scenario when training and testing data are close to each other. We first use the measurements from a five-meter segment on the trajectory as the training data to fit a friction coefficient. Then we apply this friction coefficient value to the next five-meter segment to predict the energy consumption. The results are plotted in the right subfigure of Fig. 8.

A few remarks are in order from this comparison experiment. (1) When the training and testing data are from separate areas (left one-third and right two-thirds), our network outperforms this baseline method by more than 10% based on the prediction error. It shows that our method has a better generalization capability, since we can estimate the energy consumption in a new environment (Fig. 5, Fig. 7) without taking extra measurements. (2) We do not use Gaussian Process (GP) to estimate the friction coefficients for testing environments as the work in [18]. When the training (available measurements) and testing data are from separate areas, we do not have any measurements in the testing area. The GP-based method needs to use the training data for the prior mean for parameters such as the friction coefficient, which is similar to what we did to obtain results in the left subfigure of Fig. 8. The work in [18] also showed that as the available measurements get sparse (further from the testing area), the estimation performance is negatively affected. (3) We do see that the prediction in the left subfigure of Fig. 8 has the same increasing/decreasing trend as the ground truth, but the scale is not accurate. It indicates that the bumpy movement of the ground robot on rough surfaces cannot be easily modeled with friction and gravity force based on physics laws. Using a learning-based method can avoid the modeling difficulty.

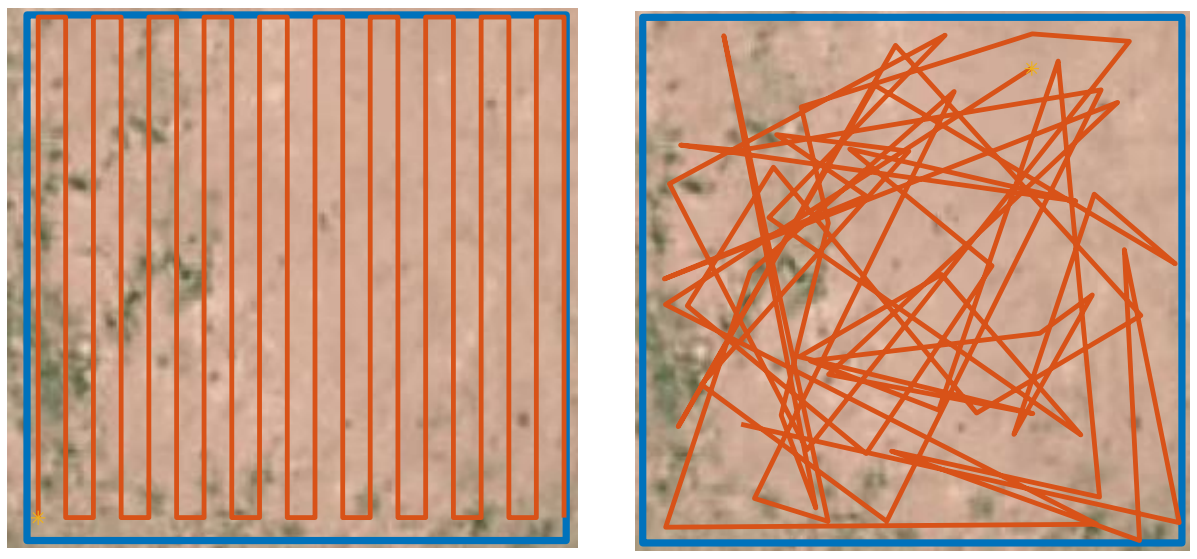

Fig. 9: We collect the measurements on a playground of a dirt area. The terrain class is different than the one for training. We use the new measurements to examine whether the trained network on grass can be useful for predicting the energy consumption on dirt.

### E. Test Results on a New Terrain Class

So far we have tested our method in two environments of the same terrain class (grass). In this subsection, we demonstrate that the trained network can still be useful for energy-efficient path planning on another terrain class that is not included in the training.

The energy consumption of a ground robot on various terrain classes can be significantly different due to the factors such as surface materials, vegetation, and wetness. It is difficult to have access to the energy measurements on all possible terrain classes to train a network in advance. Furthermore, even for the same terrain class, the energy consumption can vary a lot. For example, the grass in the wild usually costs more energy for a ground robot compared to a well-maintained grass terrain on a golf course. Therefore, it is uneasy to train a network that can have high accuracy for all environments.

We collect the measurements on a playground of a dirt area with both back-and-forth and random-waypoint motion patterns, as shown in Fig. 9. The ground surface of this area is smoother than the grass surface for training, which makes it more energy-efficient for navigation. The average energy consumption on this dirt terrain is lower than the previous test environment. Therefore, directly applying the network trained in Sec. IV leads to a relatively large prediction error of 19.6%, as shown in the left subfigure of Fig. 10.

However, though the absolute energy prediction values deviate significantly from the ground truth, we can see that the relative scale is still close. More specifically, let $\hat{e}$ denote the prediction of our network, $e^*$ denote the ground truth, and $\hat{e}' = \hat{e} - mean(\hat{e}) + mean(e^*)$. $\hat{e}'$ has the same mean as $e^*$. In the right subfigure of Fig. 10, we plot $\hat{e}'$ (green) and $e^*$ (red) together. The average error is 12.6%, which is comparable to the test results on the grass areas.

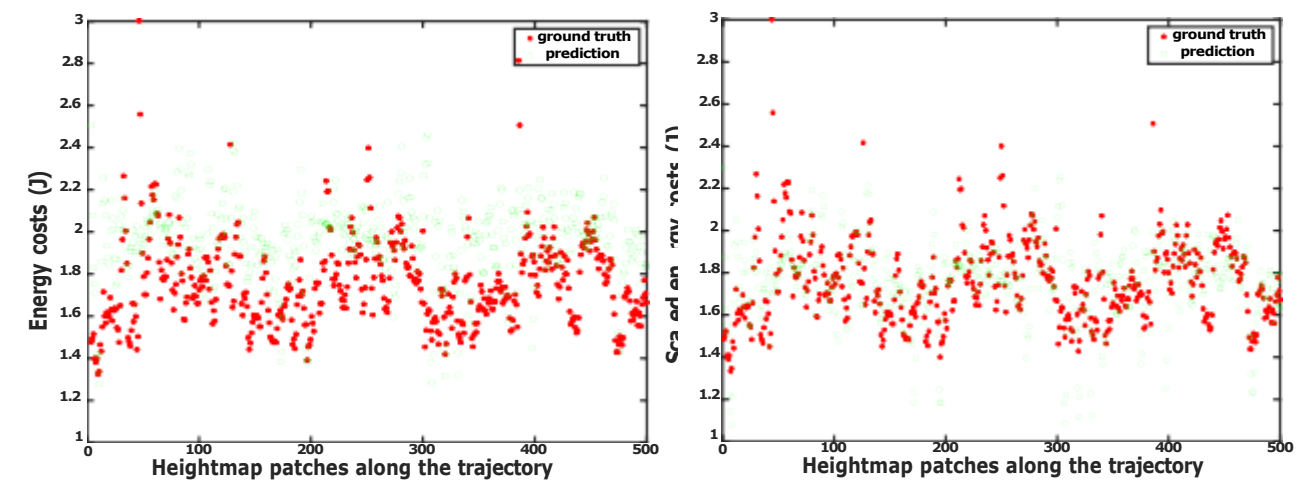

Fig. 10: Part of the test results on the test dirt terrain. This area is more energy-efficient for navigation compared to the grass area. *left.* Directly applying the pre-trained network on grass leads to a relatively large error of 19.6%. However, though the absolute value of the prediction deviates from the ground truth, the relative scale is similar. *right.* We translate the prediction to make it have the same mean as the ground truth. The error rate drops to 12.6%, which is comparable to the test results on the grass terrain. The test result can still be useful for planning the most energy-efficient path to the goal locations in this environment.

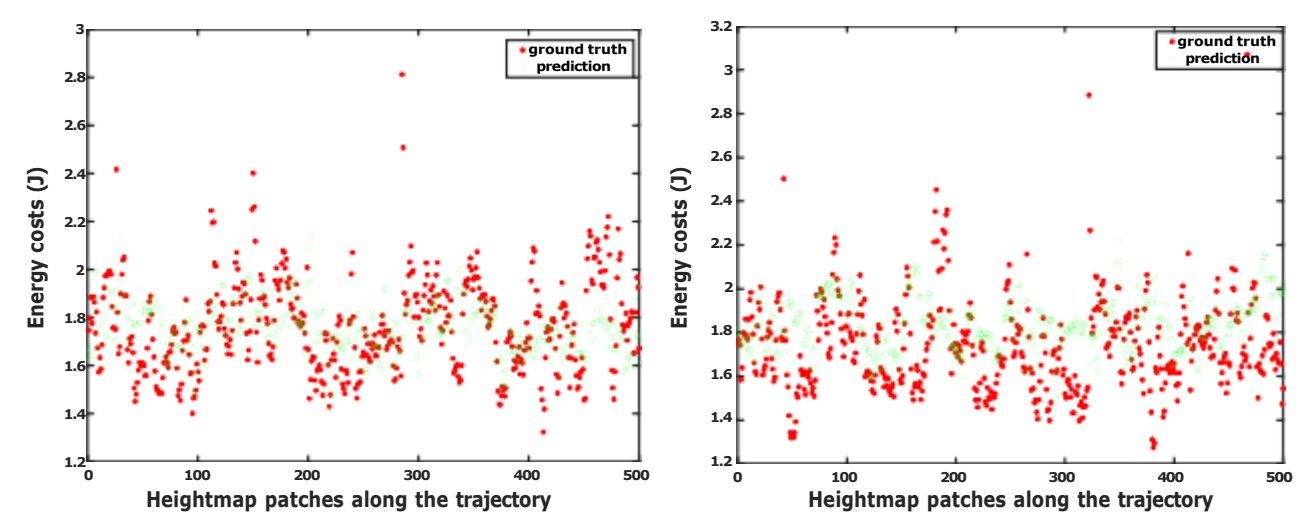

Fig. 11: Part of the test results on the new terrain class (dirt) after fine-tuning. The average error is 9.4% (left subfigure) for the path in the left subfigure of Fig. 9 and 12.2% (right subfigure) for the path in the right subfigure of Fig. 9. The performance improves and is similar to the testing results on the grass terrain after fine-tuning.

In addition, we also demonstrate that we can fine-tune the pre-trained ResNet-based network to improve the prediction performance on a new terrain class instead of training a network from scratch. For this purpose, we fine-tune the network with part of the data collected in the left subfigure of Fig. 9. Only the measurements from the left one-third area are used for fine-tuning and the rest is for evaluation. The size of the data for fine-tuning is 1000, which is close to the initial training. However, we only fine-tune the network for five minutes. The pretrained network is able to quickly adjust to the new data. The left subfigure of Fig. 11 illustrates the prediction results of the fine-tuned network. The error rate improves to 9.4%. We also evaluate the fine-tuned network for the trajectory in the right subfigure of Fig. 9. The average error is 12.2%, as shown in the right of Fig. 11. After fine-tuning, the performance improves to the same level as the tests on the grass terrain.

## VI. CONCLUSION

In this paper, we presented a method to predict the energy consumption of a ground robot for navigation in uneven, off-road environments. We showed that model-based methods using simple physical laws cannot model the energy consumption accurately on these terrains. In contrast, our method performs better and generalizes to new environments with various navigation directions and slope angles.

In our current model, the robot motion inside a patch is assumed to be at a constant speed. A future research direction is to remove this assumption and consider more general trajectories within the patch. For example, We could include the robot velocity in our model so that the energy consumption can be predicted when the robot turns and accelerates in the heightmap patch.